# The Compilation of Decision Models*

**David E. Heckerman**
Medical Computer Science Group
Knowledge Systems Laboratory
Departments of Computer Science and Medicine
Stanford, California 94305

**John S. Breese**
Rockwell International Science Center
Palo Alto Laboratory
444 High Street
Palo Alto, California 94301

**Eric J. Horvitz**
Medical Computer Science Group
Knowledge Systems Laboratory
Departments of Computer Science and Medicine
Stanford, California 94305

## Abstract

We introduce and analyze the problem of the compilation of decision models from a decision-theoretic perspective. The techniques described allow us to evaluate various configurations of compiled knowledge given the nature of evidential relationships in a domain, the utilities associated with alternative actions, the costs of run-time delays, and the costs of memory. We describe procedures for selecting a subset of the total observations available to be incorporated into a compiled situation–action mapping, in the context of a binary decision with conditional independence of evidence. The methods allow us to incrementally select the best pieces of evidence to add to the set of compiled knowledge in an engineering setting. After presenting several approaches to compilation, we exercise one of the methods to provide insight into the relationship between the distribution over weights of evidence and the preferred degree of compilation.

## 1 Introduction

There has been growing interest in reducing complex deliberation in computer-based reasoners by developing decision-making techniques that rely on precomputed or compiled responses. In particular, recent *reactive planning* research has centered on the replacement of unwieldy solution mechanisms and detailed representations of knowledge with compiled strategies that enable agents to respond, in reflex fashion, to relatively simple perceptual inputs[1,3]. It has been hoped that, for many contexts, explicit representations and deliberation will not be necessary for sufficient performance.

To date, most of the work on compilation has been done in the absence of theoretical foundations of decision making under uncertainty. The goal has been to create reasoners with satisficing behavior. However, there has been growing interest in the optimal design of problem solvers and solution methodologies through the application of normative principles [8]. In this paper, we describe methods for the efficient representation and indexing of precomputed actions from a perspective of *bounded optimality*. Bounded-optimal systems perform ideally in some context or set of contexts under expected constraints in reasoning resources, such as in time and memory [9]. We define optimality in the context of a decision-theoretic analysis of the costs and benefits of alternative hardware configurations and strategies for determining the best action. As viewed from the perspective of decision analysis, the compilation techniques to date [14,1,3,11] typically have been suboptimal as investigators have relied on ad hoc models.

*This work was supported by NASA under Grant NCC-220-51, by the NSF under Grant IRI-8703710, and by the NLM under Grant RO1LM04529.

Recent discussions on compilation, from the decision-theoretic perspective, have explored the derivation and consistency of compiled situation-action rules from underlying normative decision models [2] and the relationship of heuristics to decision-theoretic models [12]. It has been suggested that decision-theoretic approaches to the compilation of action, and the integration of compiled and computed approaches, may be useful in the development of tractable normative reasoning systems [9]. Indeed, recent work on the speed up of probabilistic inference through caching probabilistic results has been promising [7]. The development of elegant and theoretically sound techniques for compiling useful patterns of behavior undoubtedly will help enhance our understanding of the production of inexpensive, fast-acting, and well-characterized computational agents from more complex, explicit models of problem solving. Formal analyses can enable us to probe mathematically the relationships among parameters of problem solvers and domains, allowing us to glean new insights about compilation. Such insights may evade us in the context of less rigorous analyses.

We address the reactive-planning problem from a decision-theoretic perspective. Our analysis centers on the application of knowledge about (1) the nature of evidential relationships in a domain, (2) the utilities associated with alternative actions, (3) the costs of run-time delay, and (4) the costs of memory. Our model provides intuition about the nature of ideal compilation strategies as a function of these parameters. The approach is similar in spirit to recent work on the control of probabilistic inference that applies knowledge about inference progress and future belief distributions to reasoning about the expected value of computation [10]. In this work, we analyze several approaches to compilation and explore the effects that different assumptions about the distribution of evidence can be expected to have on the optimal configuration of compiled knowledge.

## 2 A Model for Diagnosis

A simple model for diagnosis is represented by the influence diagram pictured in Figure 1. In this model, the utility of a situation, represented by the diamond-shaped node, labeled $U$, depends on (1) whether a particular hypothesis $H$ is true or false, and (2) whether an action $D$ is taken or not. We cannot observe $H$ directly; rather, we are forced to do inference about the relative likelihood of $H$ by observing pieces of evidence $E_1, E_2, \ldots, E_n$. Thus, we seek to make a decision about a best action in the general case where we have incomplete information about the world. We shall simplify our analysis by considering the case where $D$, $H$, and the $E_i$'s are binary-valued variables (i.e., are true or false). To simplify our model further, we shall assume that all evidence is conditionally independent, given $H$ or $\neg H$. In Section 5, we examine how these assumptions can be relaxed. Finally, for definiteness, we assume that it is appropriate to take action $D$ (as opposed to $\neg D$) if $H$ is true.

Using the assumption of conditional independence of evidence, given the hypothesis and its negation, we can calculate the posterior probability of the hypothesis by multiplying together all of the likelihood ratios, $\frac{p(E_i|H)}{p(E_i|\neg H)}$, with the prior odds, $\frac{p(H)}{p(\neg H)}$.

$$\frac{p(H|E_i,\ldots,E_m)}{P(\neg H|E_i,\ldots,E_m)} = \frac{p(E_1|H)}{p(E_1|\neg H)} \cdots \frac{p(E_m|H)}{p(E_m|\neg H)} \frac{p(H)}{p(\neg H)}$$

This equation can be written more compactly in odds form, as

$$O(H|E_i,\ldots,E_m) = O(H)\prod_{i=1}^{m} \lambda_i \qquad (1)$$

where $\lambda_i$ is the likelihood ratio $\frac{p(E_i|H)}{p(E_i|\neg H)}$. Thus, we require $m$ multiplications for a complete update in light of $m$ pieces of evidence.

Because $D$ and $H$ are binary, we can simplify our reasoning about whether we should take action $D$ by calculating a threshold probability, $p^*$, for the decision. This threshold probability is defined as the probability $H$ at which we are indifferent between acting and not acting. That is, we define $p^*$ as the point where acting and not acting have equal utility, or

$$p^*U(H,D) + (1-p^*)U(\neg H, D) = p^*U(H, \neg D) + (1-p^*)U(\neg H, \neg D)$$

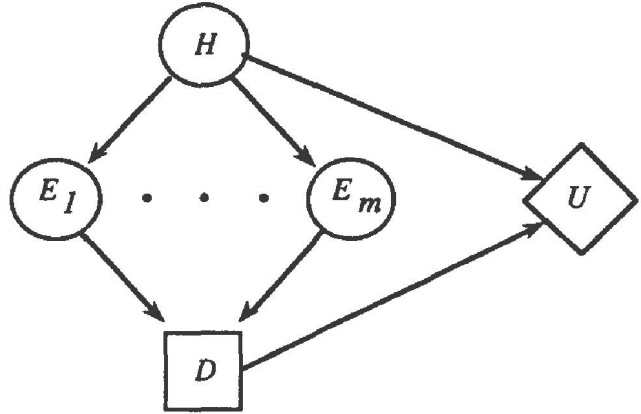

Figure 1: An influence-diagram representation of the diagnosis problem. The utility (diamond node, $U$) depends on a hypothesis (circular node, $H$) and a decision (square node, $D$). All evidence is available at the time the decision is made.

It is better to act if and only if $p(H|E_1 \ldots, E_m) > p^*$. In terms of the odds formulation, we wish to act if and only if

$$O(H|E_1, \ldots, E_m) \geq \frac{p^*}{1-p^*}$$

The weight of evidence, $w_i$, is defined as the log of the likelihood ratio, $\ln \lambda_i$. Mapping likelihood ratios into weights of evidence allows us to update belief in a probability through the addition of the weights of evidence. (See [4] for a discussion of the use of weights of evidence in the analysis of belief-updating schemes in formalisms for managing uncertainty in artificial intelligence.) We can rewrite the threshold-probability condition in terms of the log-likelihood ratio where $w_i = \ln \lambda_i$. It is better to act if and only if

$$W = \sum_{i=1}^{m} w_i \geq \ln \frac{p^*}{1-p^*} - \ln O(H) = W^* \tag{2}$$

In this expression, $W^*$ is the threshold sum of weights of evidence for the decision problem. We will use this formulation of diagnostic decision making, in conjunction with the assumption that all evidence is conditionally independent, to analyze the tradeoffs associated with computation versus compilation.

## 3 Computation Versus Compilation

Given a simple decision problem defined within the model of diagnosis and analytic framework described in Section 2, two fundamental design alternatives are:

1. **Compute.** Endow our computational decision maker with the ability to perform probabilistic inference in real-time, and commit to a decision that maximizes expected utility. That is, given a vector of information about the truth status of all of the relevant evidence, we compute

$$W = \sum_{i=1}^{m} w_i$$

the sum of the weights of evidence for all observations. We act if and only if $W > W^*$.

2. **Compile.** Select $n$ pieces of evidence,

$$\mathbf{E}_c = \{E_{c_1}, \ldots, E_{c_n}\} \subseteq \{E_1, \ldots, E_m\}$$

and calculate

$$W_n = \sum_{i=1}^{n} w_{c_i}$$

for each possible instance of the variables in $\mathbf{E}_c$. If $W_n$ exceeds $W^*$, we store $D$ in a lookup table; otherwise, we store $\neg D$. There are $2^n$ possible observations; thus, to access the compiled solutions, we require a memory cache of a size that grows exponentially with the number of pieces of evidence considered (an $n$-dimensional array).

## 3.1 Net Inferential Value

We wish to compare the relative costs and benefits of these two alternatives. To do so, we determine the *net inferential value* (NIV) of each alternative. The net inferential value of an alternative is the expected net present value of that alternative over the entire lifetime of its use. In this paper, we decompose this value into three components. The first component is the expected value of the recommendations provided by each method. In the general case, the compilation method will ignore some available information. Thus, the quality of its recommendations typically will be lower than those derived from computing. The second component is the memory requirements of computation and compilation. Finally, we must consider the cost of processing time, based in delayed response, when a case presents itself. This cost may be depend on whether $H$ is true (e.g., suppose $H$ is the hypothesis that a patient is having a heart attack).

Let $\text{EV}_{\text{compute}}$ denote the expected value of a system committed to the compute policy for a single episode. Also, let $\text{PC}^H_{\text{compute}}$ and $\text{PC}^{\neg H}_{\text{compute}}$ denote the processing costs for computation given $H$ is true and false, respectively. Finally, let $\text{MC}_{\text{compute}}$ be the the memory costs associated with the compute policy. Let $\text{EV}_{\text{compile}}$, $\text{PC}^H_{\text{compile}}$, $\text{PC}^{\neg H}_{\text{compile}}$, and $\text{MC}_{\text{compile}}$ denote similar quantities for the compile policy. It follows that the net inferential values of the compute and compile policies are

$$\text{NIV}_{\text{compute}} = r \left[ \text{EV}_{\text{compute}} - \text{PC}^H_{\text{compute}} \, p(H) - \text{PC}^{\neg H}_{\text{compute}} \, p(\neg H) \right] - \text{MC}_{\text{compute}} \tag{3}$$

$$\text{NIV}_{\text{compile}} = r \left[ \text{EV}_{\text{compile}} - \text{PC}^H_{\text{compile}} \, p(H) - \text{PC}^{\neg H}_{\text{compile}} \, p(\neg H) \right] - \text{MC}_{\text{compile}} \tag{4}$$

where $r$ is a factor that converts the expected value of each policy on a single problem instance to a summary (present) value for a series of problem instances over the life of the system. (A discussion of bounded optimality and related issues on the value of a reasoning system over a lifetime and over a specific time horizon is found in [9].) We choose to compute if and only if

$$\text{NIV}_{\text{compute}} \geq \text{NIV}_{\text{compile}} \tag{5}$$

In our simple model for inference, Equation 2, the expected value of a system committed to the compute policy for a single episode is

$$\begin{aligned}\text{EV}_{\text{compute}} &= [p(W > W^*|H) \, U(H, D) + p(W \leq W^*|H) \, U(H, \neg D)] \, p(H) + \\ &\quad [p(W > W^*|\neg H) \, U(\neg H, D) + p(W \leq W^*|\neg H) \, U(\neg H, \neg D)] \, p(\neg H)\end{aligned} \tag{6}$$

The similar expression for compilation is

$$\begin{aligned}\text{EV}_{\text{compile}}{}^n &= [p(W_n > W^*|H) \, U(H, D) + p(W_n \leq W^*|H) \, U(H, \neg D)] \, p(H) + \\ &\quad [p(W_n > W^*|\neg H) \, U(\neg H, D) + p(W_n \leq W^*|\neg H) \, U(\neg H, \neg D)] \, p(\neg H)\end{aligned} \tag{7}$$

The processing time for computation is linear in the total number of pieces of evidence. The processing time for the compile policy is also linear in the total number of pieces of evidence, but the proportionality constant is smaller. Also, for simplicity, we assume that the cost of delay is linear in the length of the delay. Thus, we obtain

$$PC_{compute}^{H} = k_1 m \qquad (8)$$
$$PC_{compute}^{\neg H} = k_2 m \qquad (9)$$
$$PC_{compile}^{H} = k_3 n \qquad (10)$$
$$PC_{compile}^{\neg H} = k_4 n \qquad (11)$$

The amount of memory required for computing is linear in the number of pieces of evidence, because we have to store $p(E_i|H)$ and $p(E_i|\neg H)$, or their likelihood ratio equivalents, for each evidence variable. The memory required for compilation is exponential in $n$, the number of evidence variables considered. Again, for simplicity, we assume that the cost associated with the use of memory is proportional to the amount of memory required. Thus,

$$MC_{compute} = k_5 m \qquad (12)$$
$$MC_{compile} = k_5 2^n \qquad (13)$$

Combining Equations 5 through 13, we choose to compute if and only if

$$r[EV_{compute} - (k_1 p(H) + k_2 p(\neg H))m] - k_5 m \geq r[EV_{compile} - (k_3 p(H) + k_4 p(\neg H))n] - k_5 2^n \qquad (14)$$

To complete the computation of net inferential value, we must evaluate $p(W > W^*|H)$ and $p(W \leq W^*|H)$. For one piece of evidence, we have

| $w_i$ | $p(w_i|H)$ | $p(w_i|\neg H)$ |
|---|---|---|
| $\ln \frac{p(E_i|H)}{p(E_i|\neg H)}$ | $p(E_i|H)$ | $p(E_i|\neg H)$ |
| $\ln \frac{p(\neg E_i|H)}{p(\neg E_i|\neg H)}$ | $p(\neg E_i|H)$ | $p(\neg E_i|\neg H)$ |

To simplify the our notation, we let $p(E_i|H) = \alpha$ and $p(E_i|\neg H) = \beta$. The expectation and variance of $w$ given $H$ and $\neg H$ is then

$$EV(w|H) = \alpha \ln \frac{\alpha}{\beta} + (1-\alpha) \ln \frac{(1-\alpha)}{(1-\beta)} \quad Var(w|H) = \alpha(1-\alpha)\ln^2 \frac{\alpha(1-\beta)}{\beta(1-\alpha)} \qquad (15)$$

$$EV(w|\neg H) = \beta \ln \frac{\alpha}{\beta} + (1-\beta) \ln \frac{(1-\alpha)}{(1-\beta)} \quad Var(w|\neg H) = \beta(1-\beta)\ln^2 \frac{\alpha(1-\beta)}{\beta(1-\alpha)} \qquad (16)$$

We can now take advantage of the additive property of weights of evidence. The central limit theorem states that sum of independent random variables approaches a normal distribution when the number of variables becomes large. Furthermore, the expectation and variance of the sum is just the sum of the expectations and variances of the individual random variables, respectively. Because we have assumed that evidence variables are independent, given $H$ or $\neg H$, the expected value of the sum of the weights of evidence $W_n$ is

$$EV(W_n|H) = \sum_{i=1}^{n} EV(w_i|H)$$

The variance of the sum of the weights is

$$Var(W_n|H) = \sum_{i=1}^{n} Var(w_i|H)$$

Thus $p(W_n|H)$, the probability distribution over $W_n$ as a function of the number of pieces of evidence, is

$$p(W_n|H) \sim N(\sum_{i=1}^{n} EV(w_i|H), \sum_{i=1}^{n} Var(w_i|H)) \qquad (17)$$

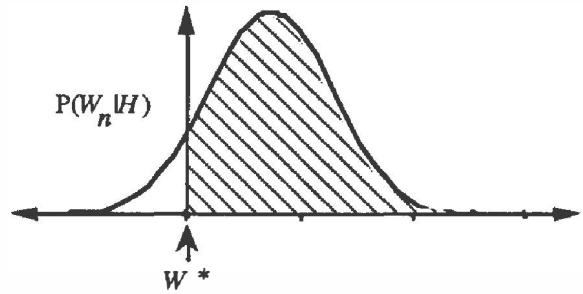

Figure 2: The probability that the total evidential weight will exceed the threshold weight is the area under the normal curve above the threshold probability (shaded region).

The equation for $\neg H$ is similar. If $n = m$ and we compile all the evidence variables, the expressions for computing and compiling are identical. Note that if $m$ or $n$ is small, making use of the central limit theorem inappropriate, we can compute the expression for the mean and variance of $W_n$ directly.

Given the distributions for $H$ and $\neg H$, we can now evaluate Equations 6 and 7 using an estimate or table of the cumulative normal distribution. We have

$$p(W_n > W^*|H) = \frac{1}{\sigma\sqrt{2\pi}} \int_{W^*}^{\infty} e^{\frac{-(t-\mu)^2}{2\sigma}} dt \qquad (18)$$

where $\mu = EV(W_n|H)$ and $\sigma = Var(W_n|H)$. The probability that the weight will exceed $W^*$ corresponds to the shaded area in Figure 2. Note that the foregoing analysis presumes that a subset of the total evidence has been selected and is available for compiling. In the next section, we address various methods for directing this selection.

## 3.2 Choice of Evidence to Compile

Let us now examine how can we select a subset of evidence, whose relevance we wish to consider for inclusion in a situation-action compilation. Each piece of evidence is associated with a pair of negative and positive weights. We would like to select those pieces of evidence that make the largest contributions to $W_n$ on an expected basis, considering their discriminatory power, measured by $\frac{p(E|H)}{p(E|\neg H)}$, and their probabilities of being observed, $p(E|H)$. Several strategies for directing the selection are possible, including (1) exhaustive search, (2) hill-climbing, and (3) stochastic simulation. Herskovits [7] discusses the third approach. In this section, we consider the second alternative.

Assume that we have already selected $n$ pieces of evidence for inclusion in the compiled set, and that we now wish to determine which remaining piece of evidence should be included in the set. In this approach, we assume that any new piece of evidence included in the set will be the *last* piece to be included. Using Equation 4, we calculate $NIV^{n+1}_{compile}$ for each prospective piece of evidence, and choose the piece that maximizes $NIV^{n+1}_{compile}$. If the value of $NIV^{n+1}_{compile}$ exceeds the value of $NIV^{n}_{compile}$, we include in the compilation set this new piece of evidence, and repeat this procedure. Otherwise, we stop searching for evidential variables to cache.

When the search procedure halts, or when additional analysis suggests that no additional evidence should be cached, we compare the net inferential value of the compilation alternative that we have generated with the net inferential value of computation. We choose to compile if and only if the former value exceeds the latter. Notice that our search procedure is heuristic; it suffers from the usual problems associated with greedy algorithms. Fortunately, standard techniques for relaxing the degree of myopia can be used to improve the approach. For example, to decrease the chance of halting at a local maximum of net inferential value, we can employ a lookahead procedure, evaluating net inferential value through one or more iterations of the algorithm, after such values begin to decline. The requirement and usefulness of various degrees of lookahead in building situation-action trees will have to be determined by analysis of real domains.

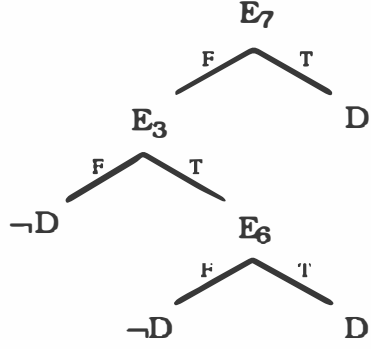

Figure 3: An asymmetrical situation-action tree. If $E_7$ is true, then we immediately act (select $D$). If $E_7$ is false, then we observe $E_3$. If $E_3$ is false, then we do not act ($\neg D$). If $E_3$ is true, then we observe $E_6$. We then act if and only if $E_6$ is true.

## 4 Alternative Compilation Strategies

In Section 3, we considered the limited set of compilation alternatives in which the appropriate decision for *every* possible instance of an evidential subset $\mathbf{E}_c$ is stored. In this section, we broaden the alternatives for compilation to include the caching of situation-action rules in asymmetrical trees. An example of such a tree, called a *situation-action tree*, is shown in Figure 3. Each path from the root to a leaf node in the tree represents a particular situation. The situations in the tree are mutually exclusive and exhaustive. The decision at each leaf in node in the tree contains the appropriate action for the situation represented by the path to that node. If evidential variable $E_7$ is true, then we act immediately (select $D$). If $E_7$ is false, then we observe $E_3$. If $E_3$ is false, then we do not act (select $\neg D$). If $E_3$ is true, then we observe $E_6$. We then act if and only if $E_6$ is true. Note that the alternatives for compilation discussed in Section 3 correspond to situation-action trees that are *symmetric*.

The computation of $\mathrm{NIV_{compile}}$ for the compilation alternative represented by the tree in Figure 3 is straightforward. If either $\{E_7\}$ or $\{\neg E_7, E_3, E_6\}$ is observed, then we select $D$. The probabilities of observing either of these mutually exclusive situations, given $H$ and $\neg H$, are

$$p(E_7|H) + p(\neg E_7|H)\, p(E_3|H)\, p(E_6|H)$$

and

$$p(E_7|\neg H) + p(\neg E_7|\neg H)\, p(E_3|\neg H)\, p(E_6|\neg H)$$

respectively. Given these expressions and similar expressions for the two paths in the situation-action tree in which $\neg D$ is prescribed, it follows that the expected value of the system's actions is

$$\begin{aligned}\mathrm{EV_{compile}} &= [p(E_7|H) + p(\neg E_7|H)\, p(E_3|H)\, p(E_6|H)]\ U(H,D)\, p(H)\, + \\ &\quad [p(E_7|\neg H) + p(\neg E_7|\neg H)\, p(E_3|\neg H)\, p(E_6|\neg H)]\ U(\neg H, D)\, p(\neg H)\, + \\ &\quad [p(\neg E_7|H)\, p(\neg E_3|H) + p(\neg E_7|H)\, p(E_3|H)\, p(\neg E_6|H)]\ U(H,\neg D)\, p(H)\, + \\ &\quad [p(\neg E_7|\neg H)\, p(\neg E_3|\neg H) + p(\neg E_7|\neg H)\, p(E_3|\neg H)\, p(\neg E_6|\neg H)]\ U(\neg H,\neg D)\, p(\neg H)\end{aligned}$$

Thus, because there are seven nodes in the tree, the net inferential value of the compilation alternative represented by the tree in Figure 3 becomes

$$\mathrm{NIV_{compile}} = r\mathrm{EV_{compile}} - 7k_5 k_6$$

where $k_6$ is cost of storing a node in an situation-action tree relative to the cost of storing the action in an array. To simplify the analysis, we have ignored the logarithmic processing cost for lookups in the tree.

In general, let $\mathcal{E}_D$ and $\mathcal{E}_{\neg D}$ be the set of all situations in a situation-action tree in which actions $D$ and $\neg D$ are prescribed, respectively. Let each situation $\vec{E}$, be represented by the set of observations in that situation. The expected value of the actions prescribed by the tree is

$$\begin{aligned}\text{EV}^{\mathcal{E}}_{\text{compile}} =\ &\left[\sum_{\vec{E}\in\mathcal{E}_D}\prod_{E\in\vec{E}}p(E|H)\right]U(H,D)\,p(H)\,+\\ &\left[\sum_{\vec{E}\in\mathcal{E}_D}\prod_{E\in\vec{E}}p(E|\neg H)\right]U(\neg H,D)\,p(\neg H)\,+\\ &\left[\sum_{\vec{E}\in\mathcal{E}_{\neg D}}\prod_{E\in\vec{E}}p(E|H)\right]U(H,\neg D)\,p(H)\,+\\ &\left[\sum_{\vec{E}\in\mathcal{E}_{\neg D}}\prod_{E\in\vec{E}}p(E|\neg H)\right]U(\neg H,\neg D)\,p(\neg H)\end{aligned} \qquad (19)$$

where $E$ refers to a particular piece of evidence (true or false) in the situation $\vec{E}$. The net inferential value of the compilation alternative associated with the tree then becomes

$$\text{NIV}^{\mathcal{E}}_{\text{compile}} = r\text{EV}^{\mathcal{E}}_{\text{compile}} - k_5 k_6 \|\mathcal{E}\| \qquad (20)$$

where $\|\mathcal{E}\|$ is the number of nodes in the tree.

Given a set of evidence variables and associated probabilities, we can use Equations 19 and 20 to direct the construction of a situation-action tree. We apply a hill-climbing search to the space of evidence variables similar to the approach described in the previous section. In this case, however, we consider extending each leaf node separately in the partially constructed situation-action tree. Specifically, we begin with the null situation-action tree in which no evidence is observed. We compute the optimal action using Equation 2, and then evaluate the net inferential value of this compilation alternative using Equations 19 and 20. We then consider each situation-action tree in which only one piece of evidence is observed. Again, we use Equation 2 to determine the appropriate action for each situation, and then use Equations 19 and 20 to evaluate the net inferential value of each situation-action tree. If none of the trees have values that are greater than the value of the null tree, we halt. Otherwise, we select the tree with the greatest net inferential value, and apply this procedure recursively to each branch of the situation-action tree.

As in the previous algorithm, when the search procedure halts, we then compare the net inferential value of the compilation alternative that we have generated with the net inferential value of computation. We choose to compile if and only if the former value exceeds the latter. As mentioned in the previous section, extending this algorithm with a lookahead procedure and other standard techniques for improving hill-climbing approaches should produce better alternatives for compilation.

The number of alternatives for compilation can be extended further. For example, we can imagine caching arbitrary sets of situation-action pairs. An efficient data structure for indexing such sets is a *trie* [7]. Two interesting problems arise with this approach. First, with the previous compilation methods, there has always been a unique cached situation that matches the observed evidence. Now, however, more than one cached situation may match the observed evidence (e.g., we cache only the two situations $\{E_1, \neg E_2\}$ and $\{E_1, E_3\}$ and we observe the evidence $E_1, \neg E_2$, and $E_3$). Approaches to handling this problem include the use of heuristic prioritization rules for making decisions about best matches and a commitment to default actions. There is also opportunity for applying decision analysis to selecting among alternative strategies for handling such incompleteness, given knowledge about the contents of the current situation-action tree. The formula for the net inferential value for such compilation alternatives of this form is similar to Equations 19 and 20. Second, the use of a greedy algortihm for identifying arbitrary sets of situation-action pairs that are valuable to cache is not as straightforward as in the previous compilation methods. A promising alternative to greedy search for selecting these pairs stochastic simulation. This approach has been explored by Herskovits [7].

# 5 Relaxation of the Assumptions

Our analysis contains several strong assumptions. Several of these assumptions can be relaxed with little effort. First, the odds-likelihood inference rule, Equation 1, and its logarithmic transform, can be extended easily to include multiple-valued evidential variables. In addition, the computation of means and variances for multiple-valued evidential variables (see Equations 15 and 16) is straightforward. Consequently, our analysis is not tied to pieces of evidence that are two-valued.

Once multiple-valued pieces of evidence are allowed, we can extend our results to inferential models in which evidence is conditionally dependent using a clustering technique described by Pearl [13] (pp. 197-204). This technique is efficient only when there are no large undirected cycles in a belief network that represents the dependencies among the evidence. There are domains such as pathology [5] in which this requirement for efficient analysis is satisfied. In some domains, however, the cycles are too large for the clustering approach to be effective [6].

Our approach can also be extended to include multiple-valued hypotheses and decisions. The algebra becomes more complex, however, because the simple $p^*$ model for action no longer applies.

We have also implicitly assumed that the cost of observing each evidential variable is zero. We can avoid making this assumption by including the cost of such observations in Equations 3, 4, and 20. In the case where asymmetric situation-action trees are an alternative for compilation, the variables observed depend on the outcome of previous observations. The algebra is extensive but straightforward.

# 6 Analysis of Prototypical Distributions

Our initial model relied on a complete characterization of the evidence used by a reasoning system in determining the probability of a proposition, and in choosing an optimal action. That is, we have assumed that we have access to the diagnostic relevance of each piece of evidence, as well as information about the probability of seeing that available evidence. In this section, we use more abstract descriptions of the patterns of evidence in a domain. In particular, we summarize the evidence in a domain by specifying a frequency distribution function over the strengths of the pieces of evidence. We can gain intuition about the compilation process by performing the analysis for some prototypical evidence patterns.

We have performed an analysis of the expected effects of compilation on the quality of decision recommendations, given the assumption that $p(E_i|H) = 1 - p(E_i|\neg H)$. This assumption imposes a symmetry in the confirmatory and disconfirmatory relevance of evidence weights. The primary ramification of the assumption of evidential symmetry is that evidence associated with higher likelihood ratios is more likely to be observed. Therefore, we can perform the summations described in the previous section by simply integrating over the pieces of evidence with the highest log-likelihood ratios, avoiding the evidence selection procedure described in Section 3.2. We have analyzed the relationship between compiling and computing for various distributions over weights, given this symmetry assumption.

Figure 4 shows a family of distributions for the weights of evidence. The curves plot the number of pieces of evidence having a particular log-likelihood ratio, $w$. A $w$ of 0 has a likelihood ratio of 1 implying $p(E|H) = p(E|\neg H) = .5$; thus, the distributions with more mass close to 0 are less informative. We have selected three linear patterns of likelihood ratios to illustrate the the effects of different levels of compilation. The distribution with more mass close to the origin indicates a model with a higher overall level of uncertainty.

Figure 5 displays Gaussian distributions of $W_n$, the sum of the compiled weights of evidence (given $H$). Each curve corresponds to a number of pieces of evidence, $n$, being included in the compiled set. The mean and variance of the distributions are calculated as in Equation 17 using the "Moderate" frequency distribution from Figure 4. Each level of compilation corresponds to constructing the compiled set of evidence by selecting those evidential variables whose log likelihood-ratios exceed a particular weight. Notice that, as we compile more pieces of evidence, the expectation and variance of the sums increase. Also, for higher levels of compilation, the probability that $W_n$ is less than $W^*$ decreases, indicating a lower probability of mistreatment as evidence is added to the compiled set.

Using these distributions, we can calculate the difference in expected value between computing and compiling for the different decay rates. The loss is the difference between $EV_{compute}$ and $EV_{compile}$, calculated using Equations 6 and 7. The expected losses in fractional terms based on compilation is shown in Figure 6 as a function of the number of pieces of evidence compiled. For the "Low" uncertainty pattern, there are several strong pieces of evidence, and therefore the losses due to compiling are small. Including about 16 elements in the compiled set results in less than a 1% reduction in performance. When uncertainty is greater, there are more substantial losses at a given level of compilation, as shown in the figure. A compilation strategy for the "High" uncertainty case which had only 1% degradation would require on the order of 50 pieces of evidence in the compiled set for a $2^{50}$ bit memory cache.

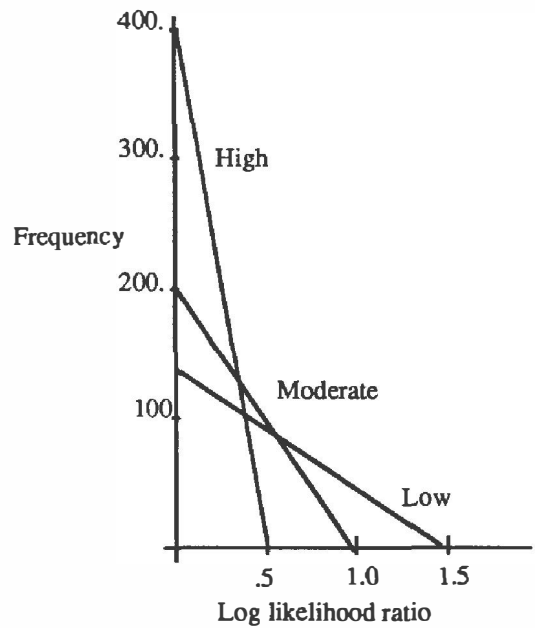

Figure 4: A family of frequency distributions over weights of evidence for various levels of uncertainty. The curve labeled "High" has weights of evidence close to zero, indicating substantial uncertainty. The "Moderate" and "Low" curves reflect lower levels of uncertainty, as measured by total evidence weights.

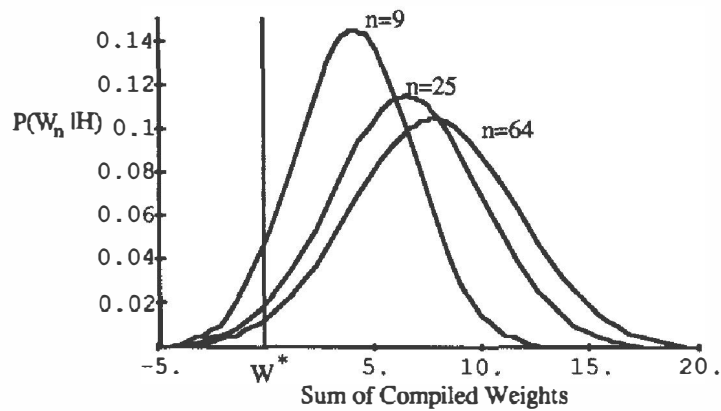

Figure 5: Probability distributions for the sum of weights of evidence (given $H$) for various levels of compilation, $n$. These distributions are based on the "Moderate" uncertainty curve.

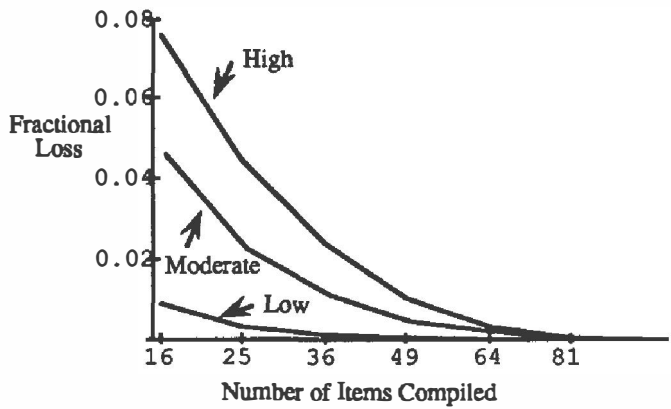

Figure 6: Fractional loss in expected value due to lower quality recommendations with compilation as a function of the number of evidence items compiled for various uncertainty profiles. As the number of pieces of evidence included in the compiled increases, the fractional loss decreases. For a given level of computation, the fractional loss is greater for higher levels of uncertainty.

Although the foregoing analysis verifies our intuition regarding the behavior of compile/compute tradeoffs for simplified decision models, there are substantial opportunities for enhancing the analysis. In particular, we plan to examine empirical data regarding the pattern of weights of evidence for actual domains to seek more realistic characterizations of prototypical distributions and to extend the analytical results.

# 7 Conclusions

We have introduced and exercised a decision-theoretic approach to trading off the costs and benefits of computation versus compilation. The method allows us to use an agent's utilities and beliefs to select the best pieces of evidence to compile at design time. We described alternative approaches to compilation, and then exercised our initial approach to gain intuition about the relationship between the distribution over weights of evidence and the ideal configuration of compilation. Although we have based our analysis on a simple decision model, many of the insights gleaned are relevant to more complex problems. Our assumptions about the binary nature of the decisions, hypotheses, and evidence, and about independence among pieces of evidence gave us model that allows for the easy computation of an optimal decision. In more complex situations, there will be greater pressures to compile. Frequently, however, there also will be greater demands on memory as the models are made more complex. There is ample opportunity to extend the analysis presented here. Our goal is to turn such analyses to evaluate alternative reactive-planning strategies in specific real-world problems.

# Acknowledgments

We thank Gregory Cooper for useful feedback. Lyn Dupre provided editorial comments on an earlier draft.